\documentclass[conference]{IEEEtran}
\IEEEoverridecommandlockouts
\usepackage[style=ieee]{biblatex}
\usepackage{amsmath,amssymb,amsfonts}
\usepackage{dirtytalk}
\usepackage[T1]{fontenc} 
\usepackage{algorithmic}
\usepackage{graphicx}
\usepackage{textcomp}
\usepackage[hidelinks]{hyperref}
\usepackage[caption=false, font=footnotesize]{subfig}
\usepackage{xcolor}
\def\BibTeX{{\rm B\kern-.05em{\sc i\kern-.025em b}\kern-.08em
    T\kern-.1667em\lower.7ex\hbox{E}\kern-.125emX}}

\usepackage{fancyhdr}
\begin{document}

\fancyhead[C]{\small \textbf{This paper has been accepted at 2023 International Joint Conference on Neural Networks (IJCNN). This is the accepted version. Please find the published version and info to cite the paper at https://doi.org/10.1109/IJCNN54540.2023.10191782}}

\title{Towards Multi-User Activity Recognition through Facilitated Training Data and Deep Learning for Human-Robot Collaboration Applications}

\author{\IEEEauthorblockN{Francesco Semeraro}
\IEEEauthorblockA{\textit{Manchester Centre for Robotics and AI} \\
\textit{The University of Manchester}\\
Manchester, UK \\
francesco.semeraro@manchester.ac.uk}
\and
\IEEEauthorblockN{Jon Carberry}
\IEEEauthorblockA{\textit{BAE Systems (Operations) Ltd.} \\
\textit{BAE Systems plc.}\\
Warton, UK \\
jon.carberry@baesystems.com}
\and
\IEEEauthorblockN{Angelo Cangelosi}
\IEEEauthorblockA{\textit{Manchester Centre for Robotics and AI} \\
\textit{The University of Manchester}\\
Manchester, UK \\
angelo.cangelosi@manchester.ac.uk}
}

\maketitle

\begin{abstract}
Human-robot interaction (HRI) research is progressively addressing multi-party scenarios, where a robot interacts with more than one human user at the same time. Conversely, research is still at an early stage for human-robot collaboration. The use of machine learning techniques to handle such type of collaboration requires data that are less feasible to produce than in a typical HRC setup. This work outlines scenarios of concurrent tasks for non-dyadic HRC applications. Based upon these concepts, this study also proposes an alternative way of gathering data regarding multi-user activity, by collecting data related to single users and merging them in post-processing, to reduce the effort involved in producing recordings of pair settings. To validate this statement, 3D skeleton poses of activity of single users were collected and merged in pairs. After this, such datapoints were used to separately train a long short-term memory (LSTM) network and a variational autoencoder (VAE) composed of spatio-temporal graph convolutional networks (STGCN) to recognise the joint activities of the pairs of people. The results showed that it is possible to make use of data collected in this way for pair HRC settings and get similar performances compared to using training data regarding groups of users recorded under the same settings, relieving from the technical difficulties involved in producing these data.

The related code and collected data are publicly available\footnote{The code repository is available at: \url{https://github.com/francescosemeraro/Multi_User_through_Single_User.git}. The data repository is available at: \url{https://figshare.com/s/64bd023e968e6eb096f3}.}.
\end{abstract}


\section{Introduction}

\label{intro}
Human-robot interaction research is constantly progressing \cite{ajoudani2018progress}. Therefore, researchers are increasingly addressing scenarios that go beyond the most commonly explored dyadic interaction between a human and a robot \cite{hornecker2022beyond,schneiders2022nonp2}. The last 15 years have been characterised by dozens of works in human-robot interaction regarding non-dyadic interactions. They mainly regard measurement of social state and engagement in crowded environments \cite{schneiders2022nona}. Despite the amount of works, they are outnumbered by the plethora of works related to HRI. Furthermore, few works tackle a human-robot collaboration (HRC) application in a multi-party scenario \cite{schneiders2022nona,schneiders2022nonp,yasar2021scalable, semeraro2023simpler}, making this a promising field for scientific investigation. Therefore, in this work, we delineate elements of a collaborative scenario in which the robot works jointly with two human users who perform concurrent tasks in a manufacturing application.

There is a growing interest also in using machine learning (ML) to build an HRC cognitive system \cite{semeraro2023human}, especially deep learning (DL) models. Indeed, in HRI it is difficult to obtain perfectly replicable experimental settings \cite{baxter2016characterising}; ML can constitute an answer to this issue, as it allows the system to be exposed, during its training, to different behaviours coming from acquisitions involving different users. Therefore, this work considers a cognitive system, based on DL models, meant to be deployed in a collaborative scenario with 2 users.

To do so, it is crucial to be able to collect data regarding the experimental scenario to train the system properly to act in the collaborative setup. However, in group interactions, this would require to have pairs of people present in the setup at the same time, introducing a further layer of complexity to the feasibility of the already demanding process of recruiting an adequate pool of participants for these studies \cite{briggs2015robots}. 

In addressing these new challenging scenarios, this work proposes a way of managing data collection of these interactions in an efficient way, that leads to promising performance in terms of learning from data.
More precisely, this study demonstrates that it is possible to build a representative training dataset regarding a non-dyadic interaction, by collecting recordings of users individually and later merging their acquisitions, to form a single data point. Achieving such result would allow non-dyadic interactions to be more easily considered in robotics research by means of machine learning. Indeed, this would leverage robotics studies from feasibility issues related to lack of data \cite{cabi2019framework} and recruitment of participants due to safety restrictions \cite{feil2020next}. From the application point of view, the study of non-dyadic HRC has the potential to unlock more complex manufacturing processes in industrial applications that go beyond the serial chain production, in which a single worker at a time can perform operations on the manufacturing target. Moreover, such result could be used for other application uses in non-dyadic HRI \cite{schneiders2022nona}, without being specific to collaborations only.

For this purpose, acquisitions related to single users were grouped in post-processing to obtain training data relatable to the robot's perception of a triadic collaborative scenario while being part of it. This allowed us to collect more data in a less demanding modality than gathering pairs of people from the beginning.

To test the performance achievable from such dataset under different configurations of DL models, a VAE and a neural network of LSTMs were trained separately to learn the joint activity of the users. We demonstrate that it is possible to achieve performances in the multi-user classification comparable to when the models are trained with data coming from users recorded at the same time and in the same setting.


The contributions of this paper are summarised below:
\begin{itemize}
    \item \textit{design outlines} of a non-dyadic HRC scenario with concurrent tasks;
    \item a \textit{multi-user activity recognition pipeline}, based on DL models trained with data coming from activities of single users and previous design considerations. 
\end{itemize}
From this point on, data generated by pairing acquisitions of single users will be addressed to as \say{\textit{grouped}}, while the ones regarding pairs of them in the same scene at the same time will be referred to as \say{\textit{pair}}.

This paper is structured as follows. Section \ref{rel_works} contextualises this work in the two main fields it contributes to. Section \ref{non-dyadic} depicts the design concepts of concurrent tasks, inside and outside HRC. Section \ref{methods} provides a breakdown of the methodology adopted to acquire and learn from the acquired data. Section \ref{res} reports the results of the trainings, then discussed in Section \ref{discussions}. Finally, Section \ref{concl} sums up the main achievements of this work.


\section{Related work}

Because of the dual aim of this paper, recent trends in the development of non-dyadic HRC and the use of DL models in group activity recognition are discussed in two different paragraphs. 

\label{rel_works}

\subsection{Non-dyadic interactions in human-robot collaborations}
Exploring non-dyadic interactions in HRI is not new \cite{schneiders2022nona}. Robotic architectures have been built for the specific purpose of studying multi-party human-machine interaction \cite{al2012furhat}. However, they mainly address situations where a shared task with the user is not considered \cite{foster2012machine,foster2017automatically}. Regarding a collaboration, non-dyadic cases are mentioned in the taxonomy, but not explored in details \cite{wang2020overview,shehadeh2017hybrid,yanco2004classifying}. Some works argue about holonic cyber-physical systems in which a scene with multiple users sharing robotic resources is contemplated, but they do not illustrate the idea with a practical case \cite{wang2016combined,shehadeh2017hybrid}. Wang et al. \cite{wang2016combined} also highlight the increasing need for robotic systems to be capable of handling these cases in the future. 

More recently, Tsamis et al. \cite{tsamis2021intuitive} carried out an experiment in which a non-dyadic HRC was involved. In this work, two human workers were present in two different workbenches at the same time, working on separate tasks. The mobile manipulator present in the validation setup had to move objects from one place to another, after one of the workers had accomplished their task.

In this work, we set a case scenario of a concurrent collaborative task between two users and one robot. The proximity and intertwining of the two roles in the design (see Section \ref{exp}) make it fall under the technical definition of a human-robot collaboration \cite{matheson2019human}.

\subsection{Multi-user activity recognition through deep learning}
Cases of multi-user activity recognition solved through DL are reported in literature \cite{li2020multi}. In 2016, Ibrahim et al. \cite{ibrahim2016hierarchical} used a LSTM to infer activity of single users to then infer the pair activity through a second LSTM. Moreover, Li et al. \cite{li2017concurrent} combined convolutional neural networks (CNNs) and LSTM for the same purpose. The main reason for this approach is that CNNs have limited capabilities of incorporating time dependencies \cite{semeraro2023human}. More recently, Shu et al. \cite{shu2020host} implemented graph networks in LSTMs and processed data coming from RGB cameras.   


In \cite{alam2021palmar}, a VAE was used to perform multi-user activity recognition from point-cloud data through a transfer learning approach. Casserfelt et al. \cite{casserfelt2019investigation} performed transfer learning with CNNs, exploiting data coming from multiple cameras. Recently, a VAE has been used by Yasar et al. \cite{yasar2021scalable} for multi-user motion prediction.

In this work, a VAE, composed of STGCN layers (see Subsection \ref{VAE}), and a LSTM netowrk were trained separately by making use of 3D skeleton poses coming from a depth camera to recognise group activity. In doing so, we propose a novel way of generating training data regarding group activity, by pairing recordings related to single users to constitute datapoints representing both of them. The main reason for choosing two models widely used in literature is to test the suitability of this data collection process to train different kinds of DL models. Indeed, the LSTM is well-suited to incorporate long-term time dependencies. Differently, the VAE may also incorporate dependencies of this kind, but it is significantly more limited in these regards. Besides, LSTMs learn through supervised learning, while VAEs perform classification through self-supervised learning (see Subsection \ref{VAE}).

\section{Design concepts for concurrent non-dyadic HRC tasks}
\label{non-dyadic}
With so few cases of HRC scenarios of this kind \cite{semeraro2023human}, it is important to outline concepts of the designed novel interaction. First, a contextualisation of concurrent non-dyadic HRC tasks is provided. Then, a concurrent non-dyadic HRC case of interest in manufacturing is described, to motivate the setting of the experimental validation for the multi-user activity recognition system.

\subsection{Concurrency in collaborative tasks}
In collaborative tasks, two agents cooperate to achieve a common goal. There are different possible ways in which this collaboration can unfold. In turn-taking collaborations, the agents mutually switch between activity and inactivity while working in the shared workspace (e.g., collaborative assembly) \cite{vinanzi2020role}. In joint tasks, they must act at the same time, due to the constraints of the task itself (e.g., lifting a heavy load together from opposite ends) \cite{roveda2020model}. However, there are concurrent tasks in which the two agents need to complete a common goal together, but jointly performing different operations, consequently causing mutual disturbances (e.g., tactical vehicle control) \cite{horrey2007examining}.

The collaborations described so far consider only two agents in the scene. The inclusion of a third agent makes sense if it performs a role different from that of the others. If the third agent acts as an entity with its own sub-task in the same common goal, thus interfering with the other agents, it could be collapsed into one of the agents \cite{yanco2004classifying}. Indeed, in this case, from the perception of one of the agents, it could be considered as a component of its counterpart in the collaboration.

A third agent could, instead, perform the role of a mediator in the collaboration. Through its intervention in the scene, it would not act on the final goal directly; rather, it would indirectly assist in the collaboration by mitigating the disturbances the two working agents may cause to each other, to improve their joint performance. The other agents would both perceive the third agent not as an extension of the counterpart, but as a different standalone agent that supports their actions. This case is the one considered in this work and is further described below.

\subsection{Multi-user activity recognition for concurrent tasks in non-dyadic human-robot collaboration}
\label{multi-user}
\begin{figure}[h]
  \centering
  \includegraphics[width=0.3\textwidth]{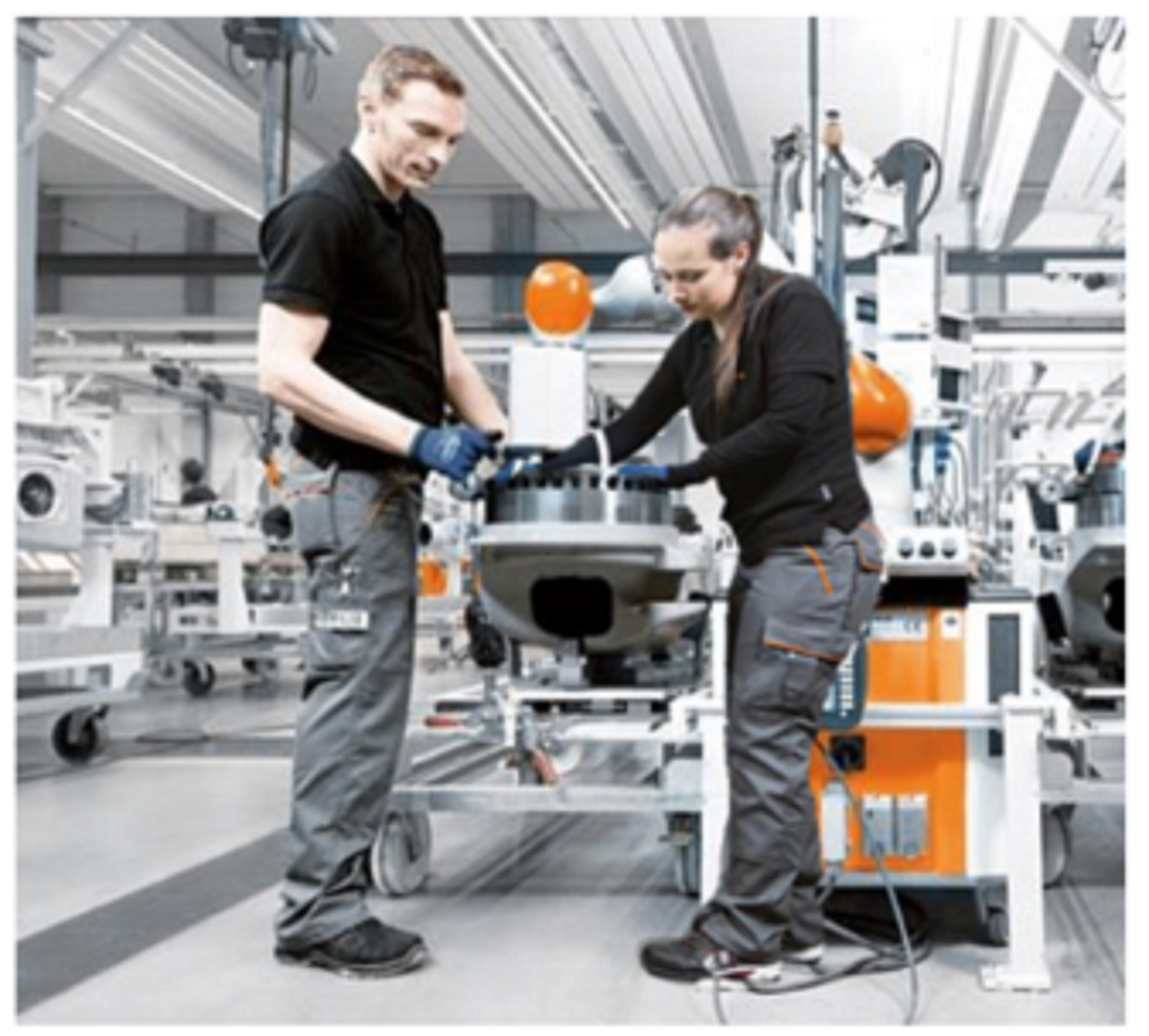}
  \caption{A collaborative robot is assisting two human users working concurrently on a manufacturing target (adapted from \cite{Quality20}).}
  \label{scenario}
\end{figure}
Based on the previous considerations, a case design of a collaboration between 2 users and 1 robot for manufacturing applications is discussed. The main reason to look at an industrial application is the recent repeatedly stated need to consider this type of scenarios in manufacturing \cite{wang2016combined}.
\begin{figure}[h] 
    \centering
  \subfloat[\label{2a}]{%
       \includegraphics[width=0.48\linewidth]{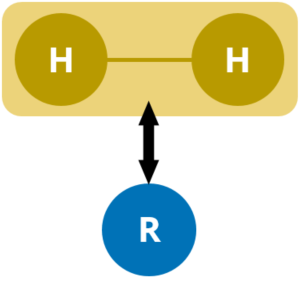}}
    \hfill
  \subfloat[\label{2b}]{%
        \includegraphics[width=0.45\linewidth]{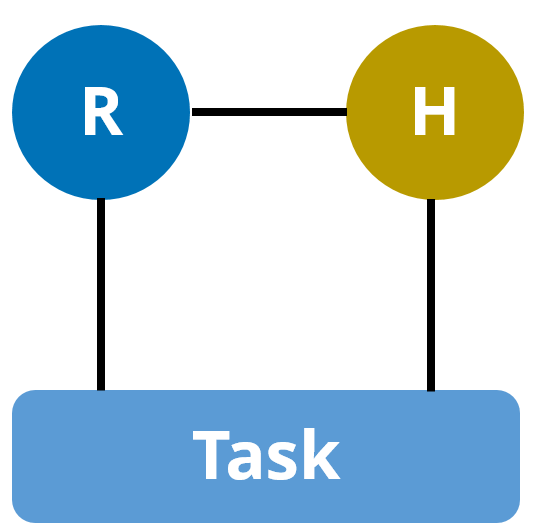}}
  \caption{Representation of the concurrent non-dyadic human-robot collaboration discussed, in terms of multiplicity (\ref{2a}) \cite{yanco2004classifying} and roles of humans and robot in relation to the task and each other (\ref{2b}) \cite{wang2017human}. In \ref{2b}, we consider the humans with an active role, while the robot with an adaptive one.}
  \label{HRC}
\end{figure}
Secondly, an industrial scenario envisions the workers performing gestures in a repetitive manner, which allows for the conditions of the setup to be more controlled.  
Nevertheless, these considerations do not exclude the potential for applications outside manufacturing, such as social robotic interactions \cite{semeraro2018physiological,fiorini2018physiological,cavallo2021mood}.

 In some production chains, one worker at a time works on the object of manufacturing. Therefore, the productivity of the process would be enhanced by enabling two workers to perform manufacturing operations on a product at the same time. However, by doing so, there is a chance that the workers may interfere with each other while performing operations on the object (e.g., by changing the position and orientation of the object to have it in an optimal configuration for the operation to perform, see Figure \ref{scenario}).  
 Such issue could be solved by the presence of a robot (e.g., a robotic manipulator endowed with vision or force sensing) that acts as an intermediary between the two users. In this way, the workers do not need to interact with each other while working, as queries regarding the state of the manufacturing target can be handled by the robot \cite{schneiders2022nonp} (see Figure \ref{HRC}).  
 
Because of this, team collaboration is replaced by the robot's mediation and the interaction between the users is reduced \cite{schneiders2022nonp}. This allows us to collect data from the scene by considering the users individually, to then treat the data coming from pairs of acquisitions as unique data points, without losing information. Being capable of acquiring data relatable to a pair setting in this way allows us to save resources, when it comes to recruiting participants.
Indeed, instead of having pairs of people at the same time in the setup, it is possible to record them individually and obtain good results in terms of multi-user activity recognition.

To pinpoint which type of activities would be crucial to recognise in such a scene, it is important to understand the role of the robot in the collaboration. More precisely, it needs to know how to handle resources of the setup (e.g., object of manufacturing) between the two users. Additionally, the robot needs to understand which user requires its attention. When the users are both working on the manufacturing, the robot needs to address requirements from both of them. When one of them finishes a sub-step of the manufacturing and prepares for the next one, the robot can shift all the resources to the user now working on its own. Finally, a user can query the robot's attention through a signal (e.g., specific hand gesture) for the robot to be aware that resources must be reallocated. The description of the unfolding of this collaboration allows us to depict three main activities the robot needs to detect from a single user:
\begin{itemize}
    \item working on the object of manufacturing;
    \item preparing for the next task;
    \item requesting a reallocation of resources from the robot.
\end{itemize}
Consequently, in terms of pair activity, 9 possible classes, related to the possible pairings of these 3 states of each user, were considered in the multi-user activity recognition problem. From this point on, these three activities will be addressed as Working (W), Preparing (P) and Requesting (R), respectively.  

\section{Materials and methods}
\label{methods}
This section delves into the technical design and validation of the multi-user activity recognition pipeline proposed. Thanks to the specified features of this novel scenario (see Subsection \ref{multi-user}), the experimental procedure built to collect the training and test data is described. Then, a breakdown of the preprocessing operations performed on the data is provided. Finally, the DL models and methods for the classification problem are explained.

\subsection{Data collection}
\label{exp}
Due to the novelty of this research, a dataset was collected to verify the proposed hypothesis that using training data coming from single users is a viable option to make inference from pair settings (see Section \ref{intro}). Not only the scenario itself is new, but also the hypothesis on the training data does not allow us to use available datasets regarding group activity \cite{ibrahim2016hierarchical}. Indeed, by disentangling such data instances to have data from single users to then merge back would not be effective to validate the hypothesis, which must use data that are comparable to the target scenario, but not exactly the same.
\begin{figure}[h] 
    \centering
  \subfloat[\label{3a}]{%
       \includegraphics[width=0.65\linewidth]{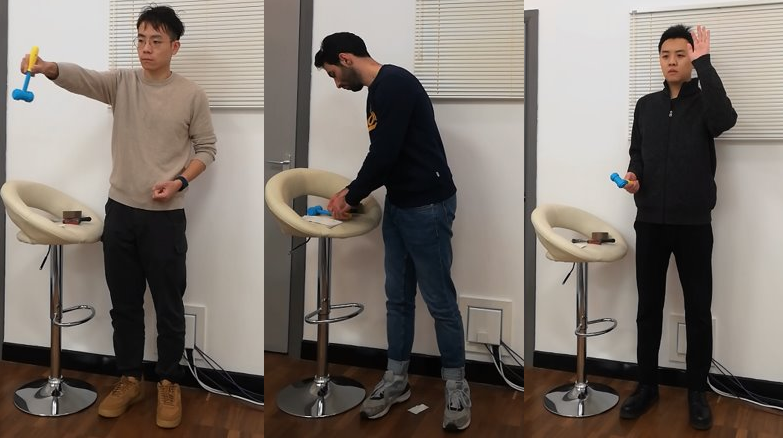}}
    \hfill
  \subfloat[\label{3b}]{%
        \includegraphics[width=0.328\linewidth]{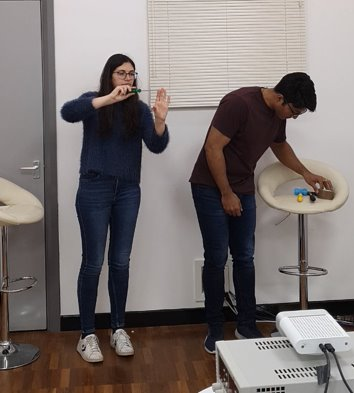}}
  \caption{Pictures from the experimental scenarios. In \ref{3a}, participants are individually performing the requested actions (from left to right: Working, Preparing and Requesting). All three of these instances will then be used to generate the grouped data. In \ref{3b}, participants are both within the field of view of the Kinect camera and are performing actions at the same time, case that corresponds to pair data.}
\end{figure}

To generate the grouped data in post-processing (see Section \ref{intro}), 11 participants were recruited for the data collection. A Microsoft® Azure Kinect camera \cite{tolgyessy2021skeleton} (resolution: 720P, frames per second: 30, depth mode: NFOV\_UNBINNED), controlled through a ROS2 Foxy workspace \cite{maruyama2016exploring} on Ubuntu 20.04 running on a Lenovo® Thinkpad P53 (Intel® Core™ i7-9750H CPU, 2.60GHz), was placed before the participants. 
With tools on a working table at their side, the participants were then individually asked to perform 3 actions, each one related to one of the aforementioned states (see Figure \ref{3a}):
\begin{itemize}
    \item performing an action relatable to the tool handled to an imaginary item of cubic shape placed at their chest height (Working);
    \item fetching an item from the working table, to prepare for the sub-task ahead (Preparing);
    \item rising a hand and hold it in place for some seconds (Requesting).
\end{itemize}

For the pair data collection, instead, 3 participants who did not participate in the grouped one were recruited. They were asked to place themselves in pairs in front of the camera controlled with the same previous settings and asked to perform the aforementioned actions in a similar way (see Figure \ref{3b}). In performing so, there was a greater level of complexity in these acquisitions. Indeed, these recordings were aimed at reproducing a scene where users work together. Therefore, instead of having many small recordings of a single action, like in the grouped data collection, longer recordings were made with the participants performing all the possible pairings of the three states. Consequently, the participants were given a sequence of grouped actions that they had to perform. These were executed naturally, without giving them any signal of switching between them. During this collection, the experimenter dynamically labeled the samples while being acquired.

\subsection{Data preprocessing}

\begin{figure*}
    \centering
    \includegraphics[width=\textwidth]{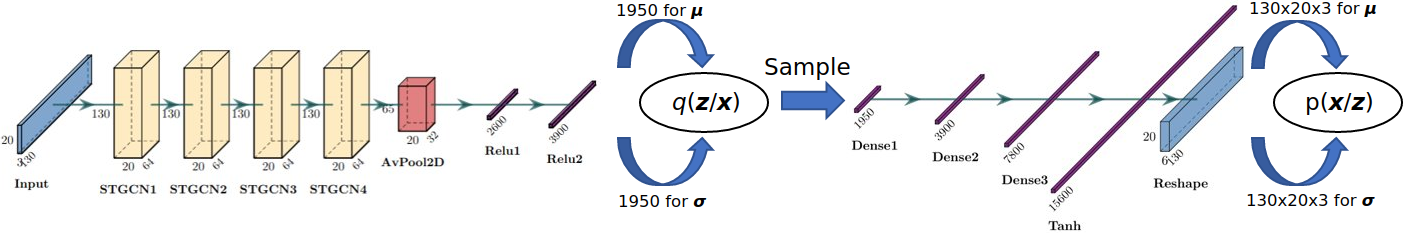}
    \caption{Structure of the VAE model used. The \textit{Reshape} layer does not contribute to the learning, but it helps visualise that the network models a distribution of the input data. The \textit{STGCN} and \textit{AvPool2D} layers were then used to perform the classification.}
    \label{VAE_model}
\end{figure*}

\begin{figure}[h] 
    \centering
  \subfloat[\label{4a}]{%
       \includegraphics[width=0.5\linewidth]{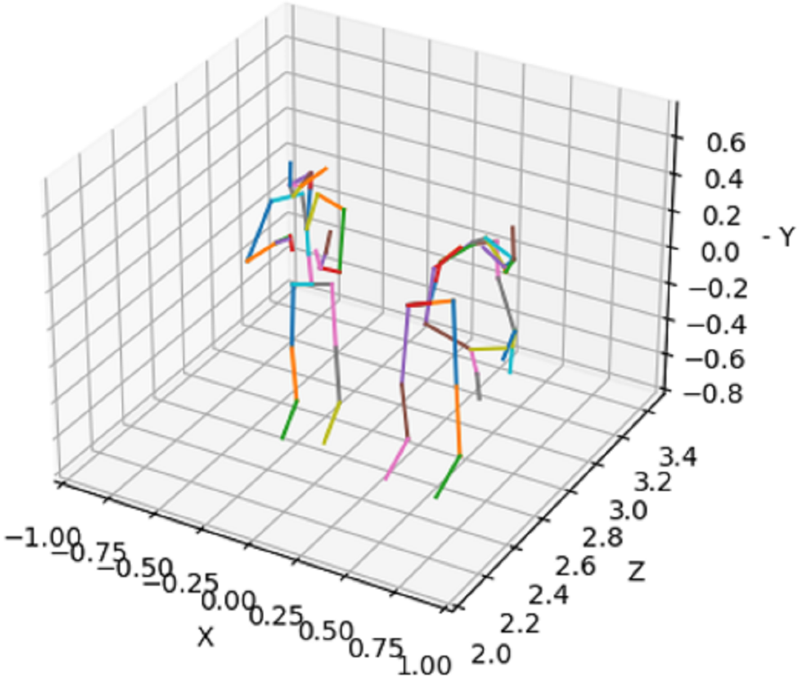}}
    \hfill
  \subfloat[\label{4b}]{%
        \includegraphics[width=0.49\linewidth]{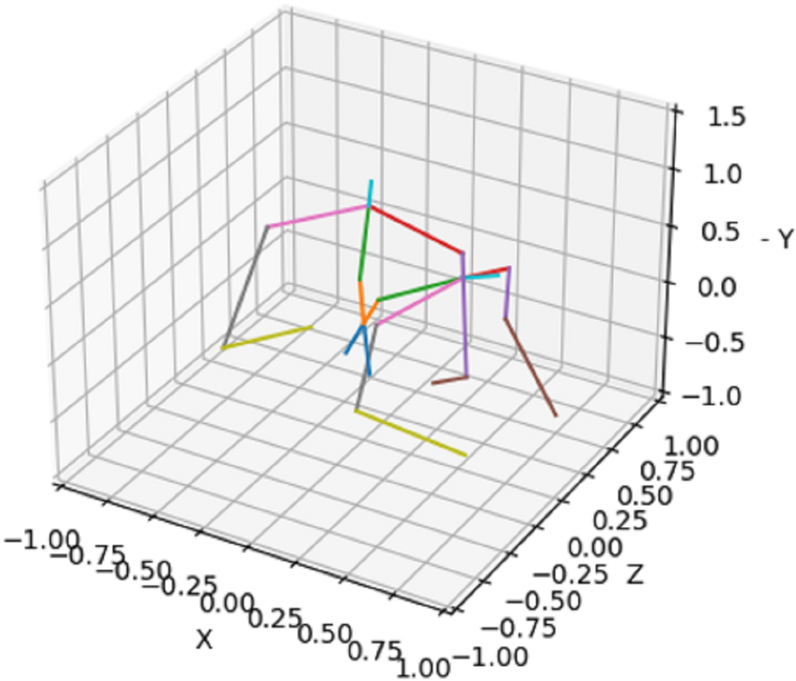}}
  \caption{Visual representation of a pair of 3D skeleton poses acquired at the same time, relatively to the Kinect camera frame, before (\ref{4a}) and after (\ref{4b}) pruning and first normalisation. In both plots, the participant on the left is working, and the one on the right is preparing for the next task.}
  \label{normalisation}
\end{figure}

Regarding the grouped data collection, the 3D skeletons underwent a preprocessing procedure. Here, 11 meaningful joints were retained from the initial set of 32 joints provided by the Kinect camera. They were the ones related to head (not including the face), neck, torso (apart from the clavicles) and arms. Then, they were normalised through the following procedure \cite{vinanzi2019mindreading}. Let us define $\textit{\textbf{J}}=[\textit{\textbf{J}}_{0}, \dots, \textit{\textbf{J}}_{10}]$ the vector of 3D coordinates of the selected joints and \textit{\textbf{J}}$_{0}$ and \textit{\textbf{J}}$_{1}$ the vector of spatial coordinates related to the spine naval and neck joints, respectively. Then, the \textit{i}-th normalised joint \textit{\textbf{J}}$_{i}$ is obtained as follows
\newline
\begin{equation}
\textbf{\textit{J}}_{i}=\frac{\textbf{\textit{J}}_{i}-\textbf{\textit{J}}_{0}}{\left\|\textbf{\textit{J}}_{1}-\textbf{\textit{J}}_{0}\right\|}
\label{form_norm}
\end{equation}
\newline
Such operation granted the 3D poses spatial invariance from the acquisition point \cite{vinanzi2019mindreading} (see Figure \ref{normalisation}).  
Furthermore, because of the formula itself, \textit{\textbf{J}}$_{0}$ will always become $(0,0,0)$, losing any informative content. Therefore, after this operation, the spine naval joint was removed. An instance of a 3D skeleton pose is consequently composed of a 10x3 vector.

\label{prep}

After this, the processed skeletons were grouped into time windows. Upon inspection of the data, it was decided to have windows of 130 samples, with 104 samples overlapping between adjacent ones, for a time window to be related to approximately 5 seconds of recordings and to introduce an approximately 1-second long new activity compared to adjacent time windows \cite{suto2020comparison,ignatov2018real}. Then, data windows coming from one user's recordings were concatenated with all the other users', generating the final grouped data points, composed of 130x20x3 coordinates (10 3D joint positions for each user in each frame). Finally, the data subdued a process of min-max normalisation, to make the range of the coordinates suitable to be an input to a neural network (see Subsection \ref{VAE}), according to the formula
\begin{equation}
\begin{split}
\textbf{\textit{J}}=&\frac{\textbf{\textit{J}}-\min (\textbf{\textit{J}})}{\max (\textbf{\textit{J}})-\min (\textbf{\textit{J}})}\times\\
\\
 \times (\operatorname{new}_{-} \max (\textbf{\textit{J}})&-\operatorname{new}_{-} \min (\textbf{\textit{J}}))+\operatorname{new}_{-}\min(\textbf{\textit{J}})\\
\\
\end{split}
\end{equation}
with $\min (\textbf{\textit{J}})=-2.5$, $\max (\textbf{\textit{J}})=1.75$, $\operatorname{new}_{-}\min(\textbf{\textit{J}})=0$ and $\operatorname{new}_{-}\max(\textbf{\textit{J}})=1$. These values were chosen by looking at the ranges of values of the grouped dataset. The same values proved to be suitable for the pair data, as well.

\begin{figure}[h]
    \centering
    \includegraphics[width=0.35\textwidth]{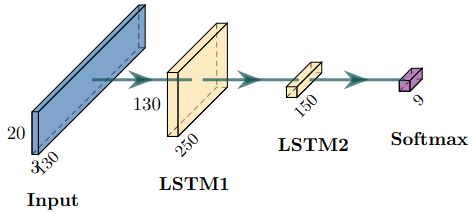}
    \caption{Structure of the cascaded LSTM network used.}
   \label{lstm}
\end{figure}
Regarding the preprocessing of the pair recordings, the normalisations were applied likewise. There was no need to pair data from different recordings, because of the experimental procedure itself (see Subsection \ref{exp}). However, it was crucial to properly discern every one of them, according to the related participant, from the data flow of skeleton frames broadcast by the camera. Indeed, because of the devised labeling (see Subsection \ref{multi-user}), the system will be able to understand not only the couple of actions being performed, but also which user is executing which. 
For instance, the cases Working-Preparing and Preparing-Working were treated as two different cases. Consequently, the data needed to be allocated properly in the 130x20x3 windows. Regarding the labeling, a data window was assigned the same label of its last pair of skeleton poses. Because the labeling was manually performed during the acquisitions (see Subsection \ref{exp}), some instances related to the transition from a couple of states to the next one could have been mislabeled. Consequently, data windows related to the 2 seconds of recording before and after a registered change of the label were discarded, to get a dataset less affected by human errors. 

\subsection{Learning framework}
\label{VAE}

To test the hypothesis on the grouped data (see Section \ref{intro}), two DL models were used. The first one is composed of stacked LSTM networks \cite{ibrahim2016hierarchical} (see Figure \ref{lstm}), to which a softmax layer of 9 neurons (related to the 9 classes to recognise, that were converted to output vectors through a one-hot encoding) was appended. This network underwent a supervised learning process by means of backpropagation through time.

The second model was a VAE \cite{alam2021palmar}. This neural network exploits a learning method related to self-supervised learning. It makes use of unlabeled data, but it uses them as labels themselves. The main objective of a VAE is to learn a variational posterior distribution $\textit{q(\textit{\textbf{z}}/\textit{\textbf{x}}})$ of latent variables \textit{\textbf{z}}, given the input data \textit{\textbf{x}}. This is done by the first half of the network, which is called encoder. Modeling this last probability distribution is done by comparing it with a prior distribution $\textit{p(\textit{\textbf{z}})}$, which in this work was a $\mathcal{N}\textit{(\textit{\textbf{0}},\textit{\textbf{I}})}$, with \textit{\textbf{0}} as vector of zeros and \textit{\textbf{I}} as identity matrix. The variational posterior then generates instances of \textit{\textbf{z}} through sampling. They are used by the second half of the network, that is the decoder, to learn a likelihood distribution $\textit{p(\textit{\textbf{x}}/\textit{\textbf{z}}})$ of the input data \textit{\textbf{x}}, given the latent variables \textit{\textbf{z}}, which is also learnt by the network. This network learns by maximizing the evidence lower bound (ELBO) loss, defined as
\newline
\begin{equation}
\textit{ELBO} = \mathbb{E}_{q_{\phi}(\textit{\textbf{z}}|\textit{\textbf{x}})} \left[\log p_{\theta}(\textit{\textbf{x}}|\textit{\textbf{z}})\right]-KL(q_{\phi}(\textit{\textbf{z}}|\textit{\textbf{x}})\|p(\textit{\textbf{z}}))
\end{equation}
\newline
where KL is the Kullback-Leibler constrastive divergence \cite{joyce2011kullback}, measure of the discrepancy between two distributions, and $\theta$ and $\phi$ are the weights of the connections of the decoder and encoder, respectively. 

The layout of the VAE is shown in Figure \ref{VAE_model} \cite{ghadirzadeh2020human}. The Spatio-Temporal Graph Convolutional Network (STGCN) layers are CNNs with adjacency matrices that are renowned for being able to learn spatial and temporal patterns of data \cite{yan2018spatial}. 
Once the training of the VAE was over, the trained STGCNs and the average pooling layer were extracted from the network and, similarly to the previously mentioned neural network, the softmax layer of 9 neurons was appended to it. This new neural network was then trained on labeled test data, keeping the weights and connections of the STGCNs unchanged. Consequently, only the softmax layer was trained during this phase. For this part of the training, a categorical crossentropy was used as the loss function. From this point on, the network composed of LSTMs will be addressed as LSTM, while the STGCNs layers trained through the VAE by means of transfer learning approach will be referred as VAE. The training and testing of both networks were performed on a container to which 2 Nvidia RTX 2080 Ti and 7 virtual CPUs (2.20 GHz), from a local server, were allocated, and in which distributed learning strategy \cite{quang2020performance} by means of Tensorflow 2 GPU and Keras libraries were used \cite{mo2017performance}.

\section{Results}
\label{res}
To evaluate whether it is possible to get valuable performances on pair data when training DL models with grouped data, the following analyses were carried out. First, training and testing by using pair data only was performed. This allowed us to get a performance baseline when testing with the pair data. Secondly, the depicted DL models were trained and tested with grouped data, to assess whether the models were suitable to learn from the grouped data. 
When the models reached a considerably good performance on this test, they then underwent the last test. In this, the networks were trained with grouped data and tested with pair data, to assess the validity of the hypothesis. Every training was performed by implementing a leave-one-subject-out (LOSO) crossvalidation. 
\begin{table}[htbp]
\caption{Accuracy and F score values in the analysed cases.}
\begin{center}
\begin{tabular}{|c|c|c|c|c|c|c|}
\hline
\textbf{Model}&\textbf{Training}&\textbf{Test}&\multicolumn{2}{|c|}{\textbf{Accuracy}}&\multicolumn{2}{|c|}{\textbf{F score}}\\
\cline{4-7} 
&\textbf{data}&\textbf{data}&\textbf{Mean}&\textbf{SD$^{\mathrm{a}}$}&\textbf{Mean}&\textbf{SD}\\
\hline
LSTM&Pair&Pair&0.548&0.139&0.676&0.116 \\
\hline
LSTM&Grouped&Grouped&0.786&0.085&0.776&0.094 \\
\hline
LSTM&Grouped&Pair&0.615&0.111&0.584&0.115 \\
\hline
VAE&Pair&Pair&0.614&0.114&0.586&0.119 \\
\hline
VAE&Grouped&Grouped&0.864&0.110&0.996&0.002 \\
\hline
VAE&Grouped&Pair&0.616&0.029&0.594&0.027 \\
\hline
\hline
\multicolumn{2}{l}{$^{\mathrm{a}}$Standard deviation} \\
\end{tabular}
\label{tab1}
\end{center}
\end{table}
\begin{figure}[h] 
    \centering
  \subfloat[\label{7a}]{%
       \includegraphics[width=0.866\linewidth]{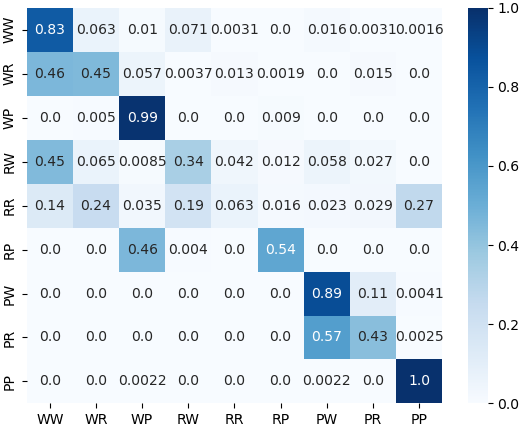}}
    \hfill
  \subfloat[\label{7b}]{%
        \includegraphics[width=0.866\linewidth]{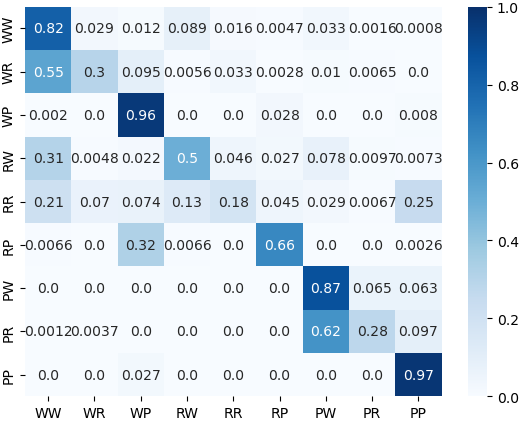}}
  \caption{Confusion matrices related to the classification performance of the LSTM (\ref{7a}) and VAE (\ref{7b}) when trained with grouped data and tested with pair data. The ground truth is plotted along the lines of the matrices, while the output of the classifier along the columns. The ratio values were obtained by dividing the number of entries in a cell by the sum of entries of the related line.}
  \label{only_test}
\end{figure}
The overall performances across all training instances are reported in Table \ref{tab1}. The reported values always refer to the test performed with the specified test data. In case the fields \say{\textit{Training data}} and \say{\textit{Test data}} have the same entry, the test set is the fold at the different iterations of the LOSO crossvalidation. Otherwise, it is the whole dataset related to that denomination (see Section \ref{intro}).
As can be appreciated in Table \ref{tab1}, the performance of the networks, when trained with grouped data and tested with pair data, is comparable to the cases in which they were trained and tested with pair data.
Therefore, as a second-level analysis, the specific performance of the DL models for each class was also surveyed. This allowed us to collect more detailed information about the whole procedure described, to find room for future improvements. Figure \ref{only_test} reports the confusion matrices of the LSTM and VAE models, using grouped data for the training and pair data for the test.

\section{Discussions}
\label{discussions}
The achieved results confirm that it is possible to make use of data collected from single users to infer related group activity without having loss in performance. Indeed, as can be seen in Table \ref{tab1}, in the VAE case with grouped data used in the training and pair data in the test, the performance of the classifiers is similar to the performance achieved by the same DL model trained and tested with pair data, both in accuracy and F scores. In the LSTM case, the accuracy value with the use of pair data only during testing is slightly higher than when the pair data were used both for training and testing, although this is evened out by the slightly lower value of F score. Besides, looking at the standard deviation values of accuracy and F scores, it is possible to point out that the VAE model generally showed more stable performances than the LSTM model. In the cases just considered, the F score values are not considerably high. Despite this, the models were still proven to be able to learn from the provided training data. Indeed, once provided with grouped data for training and testing, both models are able to achieve good performances, followed by high F scores. 



This demonstrates that it is feasible to train DL models for multi-user activity recognition, using data from single users grouped at the post-processing stage – which significantly decreases feasibility issues related to safety restrictions \cite{feil2020next} and recruiting participants for non-dyadic HRC studies \cite{cabi2019framework}. The performance of this multi-user activity recognition pipeline has proven successful for the purpose of verifying the hypothesis on grouped data, but it still has large room for improvement. First, the results showed higher reliability on the pair data than on the grouped ones. As such, future related works and experimental settings would benefit from the development of new procedures to acquire data that would strengthen the already promising results achieved in this research from this perspective. As can be observed in Figure \ref{only_test}, distinguishing between the Requesting and Working states has constituted one of the main challenges for both DL models. This appears to be mostly due to the hand position in the Requesting state, which may be confused with that of Working. Future works could use more diverse gestures for these states, in order to improve the performance of the classifier. Moreover, further defining the characteristics of the interaction would also simplify the classification task. 
Indeed, this work has considered all possible cases of a concurrent non-dyadic HRC. However, some of them could not happen in a specific study case, hence they may not be worth learning, decreasing the complexity of the situations the classifier is exposed to.

\section{Conclusions}
\label{concl}
This work has delineated outlines of experimental design for concurrent collaborations and depicted them in the case of non-dyadic human-robot collaborations. For these scenarios, this research proposed a novel way to collect data about pairs of people for training a multi-agent activity recognition system based on deep learning models. In particular, training data was realised by collecting users' activities individually and pairing them in post-processing. These data were used to separately train a long short-term memory network and a variational autoencoder to classify activities among 9 possible states of the pair of workers. The layout and training method of the networks differed from each other to test the suitability of different models to the classification problem. Both were trained with data related to grouped settings and proved not to lose in performance when tested on pair data, compared to when trained uniquely with datapoints from a pair setting. This result is promising and shows that it is possible to learn from data acquired in this way, relieving from the organisational burden of setting up multi-party scenarios for these applications.


\section*{Acknowledgments}
Francesco Semeraro’s work was supported by the DTP CASE Conversion award "Human-Robot Collaboration for Flexible Manufacturing" (Ref. 2480772), funded by UKRI Engineering and Physical Sciences Research Council and BAE Systems Plc. Angelo Cangelosi’s work was partially supported by the H2020 project TRAINCREASE (Ref. 210638710), the UKRI Trustworthy Autonomous Systems Node on Trust (Ref. EP/V026682/1) and the US Air Force project THRIVE++ (Ref. FA9550-19-1-7002).

\printbibliography

@inproceedings{semeraro2023simpler,
  title={Simpler Rather than Challenging: Design of Non-Dyadic Human-Robot Collaboration to Mediate Human-Human Concurrent Tasks},
  author={Semeraro, Francesco and Carberry, Jon and Cangelosi, Angelo},
  booktitle={Proceedings of the 22nd International Conference on Autonomous Agents and Multiagent Systems},
  year={2023}
}

@inproceedings{ibrahim2016hierarchical,
  title={A hierarchical deep temporal model for group activity recognition},
  author={Ibrahim, Mostafa S and Muralidharan, Srikanth and Deng, Zhiwei and Vahdat, Arash and Mori, Greg},
  booktitle={Proceedings of the IEEE conference on computer vision and pattern recognition},
  pages={1971--1980},
  year={2016}
}

@article{li2017concurrent,
  title={Concurrent activity recognition with multimodal CNN-LSTM structure},
  author={Li, Xinyu and Zhang, Yanyi and Zhang, Jianyu and Chen, Shuhong and Marsic, Ivan and Farneth, Richard A and Burd, Randall S},
  journal={arXiv preprint arXiv:1702.01638},
  year={2017}
}

@article{li2020multi,
  title={Multi-user activity recognition: Challenges and opportunities},
  author={Li, Qimeng and Gravina, Raffaele and Li, Ye and Alsamhi, Saeed H and Sun, Fangmin and Fortino, Giancarlo},
  journal={Information Fusion},
  volume={63},
  pages={121--135},
  year={2020},
  publisher={Elsevier}
}

@article{shu2020host,
  title={Host--parasite: Graph LSTM-in-LSTM for group activity recognition},
  author={Shu, Xiangbo and Zhang, Liyan and Sun, Yunlian and Tang, Jinhui},
  journal={IEEE transactions on neural networks and learning systems},
  volume={32},
  number={2},
  pages={663--674},
  year={2020},
  publisher={IEEE}
}

@misc{feil2020next,
  title={Where to next? The impact of COVID-19 on human-robot interaction research},
  author={Feil-Seifer, David and Haring, Kerstin S and Rossi, Silvia and Wagner, Alan R and Williams, Tom},
  journal={ACM Transactions on Human-Robot Interaction (THRI)},
  volume={10},
  number={1},
  pages={1--7},
  year={2020},
  publisher={ACM New York, NY, USA}
}

@misc{Quality20,
  author        = "Corey Ryan",
  year          = "2020",
  title         = "The Fine Line between Industrial and Collaborative Robots: It’s Smaller than You Think",
  howpublished  = "Article",
  url           = "https://www.qualitymag.com/articles/96119-the-fine-line-between-industrial-and-collaborative-robots-its-smaller-than-you-think",
  day           = 27,
  month         = jun,
  lastaccessed  = "October 5, 2022",
  note          =  ""
}

@inproceedings{casserfelt2019investigation,
  title={An investigation of transfer learning for deep architectures in group activity recognition},
  author={Casserfelt, Karl and Mihailescu, Radu-Casian},
  booktitle={2019 IEEE International Conference on Pervasive Computing and Communications Workshops (PerCom Workshops)},
  pages={58--64},
  year={2019},
  organization={IEEE}
}

@inproceedings{alam2021palmar,
  title={Palmar: Towards adaptive multi-inhabitant activity recognition in point-cloud technology},
  author={Alam, Mohammad Arif Ul and Rahman, Md Mahmudur and Widberg, Jared Q},
  booktitle={IEEE INFOCOM 2021-IEEE Conference on Computer Communications},
  pages={1--10},
  year={2021},
  organization={IEEE}
}

@inproceedings{shehadeh2017hybrid,
  title={Hybrid teams of industry 4.0: A work place considering robots as key players},
  author={Shehadeh, Mohammad A and Schroeder, Stefan and Richert, Anja and Jeschke, Sabina},
  booktitle={2017 IEEE International Conference on Systems, Man, and Cybernetics (SMC)},
  pages={1208--1213},
  year={2017},
  organization={IEEE}
}

@incollection{al2012furhat,
  title={Furhat: a back-projected human-like robot head for multiparty human-machine interaction},
  author={Al Moubayed, Samer and Beskow, Jonas and Skantze, Gabriel and Granstr{\"o}m, Bj{\"o}rn},
  booktitle={Cognitive behavioural systems},
  pages={114--130},
  year={2012},
  publisher={Springer}
}

@article{wang2016combined,
  title={Combined strength of holons, agents and function blocks in cyber-physical systems},
  author={Wang, Lihui and Haghighi, Azadeh},
  journal={Journal of manufacturing systems},
  volume={40},
  pages={25--34},
  year={2016},
  publisher={Elsevier}
}

@inproceedings{mo2017performance,
  title={Performance of deep learning computation with TensorFlow software library in GPU-capable multi-core computing platforms},
  author={Mo, Young Jong and Kim, Joongheon and Kim, Jong-Kook and Mohaisen, Aziz and Lee, Woojoo},
  booktitle={2017 Ninth International Conference on Ubiquitous and Future Networks (ICUFN)},
  pages={240--242},
  year={2017},
  organization={IEEE}
}

@inproceedings{quang2020performance,
  title={Performance evaluation of distributed training in Tensorflow 2},
  author={Quang-Hung, Nguyen and Doan, Hieu and Thoai, Nam},
  booktitle={2020 International Conference on Advanced Computing and Applications (ACOMP)},
  pages={155--159},
  year={2020},
  organization={IEEE}
}

@inproceedings{maruyama2016exploring,
  title={Exploring the performance of ROS2},
  author={Maruyama, Yuya and Kato, Shinpei and Azumi, Takuya},
  booktitle={Proceedings of the 13th International Conference on Embedded Software},
  pages={1--10},
  year={2016}
}

@article{ajoudani2018progress,
  title={Progress and prospects of the human--robot collaboration},
  author={Ajoudani, Arash and Zanchettin, Andrea Maria and Ivaldi, Serena and Albu-Sch{\"a}ffer, Alin and Kosuge, Kazuhiro and Khatib, Oussama},
  journal={Autonomous Robots},
  volume={42},
  number={5},
  pages={957--975},
  year={2018},
  publisher={Springer}
}

@article{wang2017human,
  title={Human--robot collaborative assembly in cyber-physical production: Classification framework and implementation},
  author={Wang, Xi Vincent and Kem{\'e}ny, Zsolt and V{\'a}ncza, J{\'o}zsef and Wang, Lihui},
  journal={CIRP annals},
  volume={66},
  number={1},
  pages={5--8},
  year={2017},
  publisher={Elsevier}
}

@inproceedings{semeraro2018physiological,
  title={Physiological wireless sensor network for the detection of human moods to enhance human-robot interaction},
  author={Semeraro, Francesco and Fiorini, Laura and Betti, Stefano and Mancioppi, Gianmaria and Santarelli, Luca and Cavallo, Filippo},
  booktitle={Italian Forum of Ambient Assisted Living},
  pages={361--376},
  year={2018},
  organization={Springer}
}

@article{matheson2019human,
  title={Human--robot collaboration in manufacturing applications: A review},
  author={Matheson, Eloise and Minto, Riccardo and Zampieri, Emanuele GG and Faccio, Maurizio and Rosati, Giulio},
  journal={Robotics},
  volume={8},
  number={4},
  pages={100},
  year={2019},
  publisher={MDPI}
}

@article{yasar2021scalable,
  title={A scalable approach to predict multi-agent motion for human-robot collaboration},
  author={Yasar, Mohammad Samin and Iqbal, Tariq},
  journal={IEEE Robotics and Automation Letters},
  volume={6},
  number={2},
  pages={1686--1693},
  year={2021},
  publisher={IEEE}
}

@inproceedings{tsamis2021intuitive,
  title={Intuitive and Safe Interaction in Multi-User Human Robot Collaboration Environments through Augmented Reality Displays},
  author={Tsamis, Georgios and Chantziaras, Georgios and Giakoumis, Dimitrios and Kostavelis, Ioannis and Kargakos, Andreas and Tsakiris, Athanasios and Tzovaras, Dimitrios},
  booktitle={2021 30th IEEE International Conference on Robot \& Human Interactive Communication (RO-MAN)},
  pages={520--526},
  year={2021},
  organization={IEEE}
}

@article{cabi2019framework,
  title={A framework for data-driven robotics},
  author={Cabi, Serkan and Colmenarejo, Sergio G{\'o}mez and Novikov, Alexander and Konyushkova, Ksenia and Reed, Scott and Jeong, Rae and Zolna, Konrad and Aytar, Yusuf and Budden, David and Vecerik, Mel and others},
  journal={arXiv preprint arXiv:1909.12200},
  year={2019}
}

@inproceedings{briggs2015robots,
  title={Are Robots Ready for Administering Health Status Surveys' First Results from an HRI Study with Subjects with Parkinson's Disease},
  author={Briggs, Priscilla and Scheutz, Matthias and Tickle-Degnen, Linda},
  booktitle={Proceedings of the tenth annual acm/ieee international conference on human-robot interaction},
  pages={327--334},
  year={2015}
}

@inproceedings{baxter2016characterising,
  title={From characterising three years of HRI to methodology and reporting recommendations},
  author={Baxter, Paul and Kennedy, James and Senft, Emmanuel and Lemaignan, Severin and Belpaeme, Tony},
  booktitle={2016 11th ACM/IEEE International Conference on Human-Robot Interaction (HRI)},
  pages={391--398},
  year={2016},
  organization={IEEE}
}

@inproceedings{vinanzi2019mindreading,
  title={Mindreading for robots: Predicting intentions via dynamical clustering of human postures},
  author={Vinanzi, Samuele and Goerick, Christian and Cangelosi, Angelo},
  booktitle={2019 Joint IEEE 9th International Conference on Development and Learning and Epigenetic Robotics (ICDL-EpiRob)},
  pages={272--277},
  year={2019},
  organization={IEEE}
}

@article{hornecker2022beyond,
  title={Beyond dyadic HRI: building robots for society},
  author={Hornecker, Eva and Krummheuer, Antonia and Bischof, Andreas and Rehm, Matthias},
  journal={interactions},
  volume={29},
  number={3},
  pages={48--53},
  year={2022},
  publisher={ACM New York, NY, USA}
}

@article{schneiders2022nona,
  title={Non-dyadic interaction: A literature review of 15 years of human-robot interaction conference publications},
  author={Schneiders, Eike and Cheon, EunJeong and Kjeldskov, Jesper and Rehm, Matthias and Skov, Mikael B},
  journal={ACM Transactions on Human-Robot Interaction (THRI)},
  volume={11},
  number={2},
  pages={1--32},
  year={2022},
  publisher={ACM New York, NY}
}

@article{ignatov2018real,
  title={Real-time human activity recognition from accelerometer data using Convolutional Neural Networks},
  author={Ignatov, Andrey},
  journal={Applied Soft Computing},
  volume={62},
  pages={915--922},
  year={2018},
  publisher={Elsevier}
}

@article{suto2020comparison,
  title={Comparison of offline and real-time human activity recognition results using machine learning techniques},
  author={Suto, Jozsef and Oniga, Stefan and Lung, Claudiu and Orha, Ioan},
  journal={Neural computing and applications},
  volume={32},
  number={20},
  pages={15673--15686},
  year={2020},
  publisher={Springer}
}

@inproceedings{schneiders2022nonp2,
  title={Non-Dyadic Human-Robot Interaction: Concepts and Interaction Techniques},
  author={Schneiders, Eike},
  booktitle={HRI'22: ACM/IEEE International Conference on Human-Robot Interaction},
  pages={1176--1178},
  year={2022},
  organization={IEEE Press}
}

@inproceedings{foster2012machine,
  title={Machine learning of social states and skills for multi-party human-robot interaction},
  author={Foster, Mary Ellen and Keizer, Simon and Wang, Zhuoran and Lemon, Oliver},
  booktitle={Proceedings of the workshop on Machine Learning for Interactive Systems (MLIS 2012)},
  pages={9},
  year={2012}
}

@article{foster2017automatically,
  title={Automatically classifying user engagement for dynamic multi-party human--robot interaction},
  author={Foster, Mary Ellen and Gaschler, Andre and Giuliani, Manuel},
  journal={International Journal of Social Robotics},
  volume={9},
  number={5},
  pages={659--674},
  year={2017},
  publisher={Springer}
}

@article{horrey2007examining,
  title={Examining cognitive interference and adaptive safety behaviours in tactical vehicle control},
  author={Horrey, William J and Simons, Daniel J},
  journal={Ergonomics},
  volume={50},
  number={8},
  pages={1340--1350},
  year={2007},
  publisher={Taylor \& Francis}
}

@article{roveda2020model,
  title={Model-based reinforcement learning variable impedance control for human-robot collaboration},
  author={Roveda, Loris and Maskani, Jeyhoon and Franceschi, Paolo and Abdi, Arash and Braghin, Francesco and Molinari Tosatti, Lorenzo and Pedrocchi, Nicola},
  journal={Journal of Intelligent \& Robotic Systems},
  volume={100},
  number={2},
  pages={417--433},
  year={2020},
  publisher={Springer}
}

@inproceedings{vinanzi2020role,
  title={The Role of Social Cues for Goal Disambiguation in Human-Robot Cooperation},
  author={Vinanzi, Samuele and Cangelosi, Angelo and Goerick, Christian},
  booktitle={2020 29th IEEE International Conference on Robot and Human Interactive Communication (RO-MAN)},
  pages={971--977},
  year={2020},
  organization={IEEE}
}

@inproceedings{schneiders2022nonp,
  title={Non-Dyadic Entrainment for Industrial Tasks},
  author={Schneiders, Eike and Celestin, Stanley},
  booktitle={Workshop on Joint Action, Adaptation, and Entrainment in Human-Robot Interaction},
  year={2022}
}

@incollection{joyce2011kullback,
  title={Kullback-leibler divergence},
  author={Joyce, James M},
  booktitle={International encyclopedia of statistical science},
  pages={720--722},
  year={2011},
  publisher={Springer}
}

@inproceedings{yan2018spatial,
  title={Spatial temporal graph convolutional networks for skeleton-based action recognition},
  author={Yan, Sijie and Xiong, Yuanjun and Lin, Dahua},
  booktitle={Thirty-second AAAI conference on artificial intelligence},
  year={2018}
}

@inproceedings{wang2020overview,
  title={Overview of human-robot collaboration in manufacturing},
  author={Wang, Lihui and Liu, Sichao and Liu, Hongyi and Wang, Xi Vincent},
  booktitle={Proceedings of 5th international conference on the industry 4.0 model for advanced manufacturing},
  pages={15--58},
  year={2020},
  organization={Springer}
}

@inproceedings{yanco2004classifying,
  title={Classifying human-robot interaction: an updated taxonomy},
  author={Yanco, Holly A and Drury, Jill},
  booktitle={2004 IEEE international conference on systems, man and cybernetics (IEEE Cat. No. 04CH37583)},
  volume={3},
  pages={2841--2846},
  year={2004},
  organization={IEEE}
}

@article{ghadirzadeh2020human,
  title={Human-centered collaborative robots with deep reinforcement learning},
  author={Ghadirzadeh, Ali and Chen, Xi and Yin, Wenjie and Yi, Zhengrong and Bj{\"o}rkman, M{\aa}rten and Kragic, Danica},
  journal={IEEE Robotics and Automation Letters},
  volume={6},
  number={2},
  pages={566--571},
  year={2020},
  publisher={IEEE}
}

@incollection{fiorini2018physiological,
  title={Physiological sensor system for the detection of human moods towards internet of robotic things applications},
  author={Fiorini, Laura and Semeraro, Francesco and Mancioppi, Gianmaria and Betti, Stefano and Santarelli, Luca and Cavallo, Filippo},
  booktitle={New Trends in Intelligent Software Methodologies, Tools and Techniques},
  pages={967--980},
  year={2018},
  publisher={IOS Press}
}

@article{cavallo2021mood,
  title={Mood classification through physiological parameters},
  author={Cavallo, Filippo and Semeraro, Francesco and Mancioppi, Gianmaria and Betti, Stefano and Fiorini, Laura},
  journal={Journal of Ambient Intelligence and Humanized Computing},
  volume={12},
  number={4},
  pages={4471--4484},
  year={2021},
  publisher={Springer}
}

@article{semeraro2023human,
  title={Human--robot collaboration and machine learning: A systematic review of recent research},
  author={Semeraro, Francesco and Griffiths, Alexander and Cangelosi, Angelo},
  journal={Robotics and Computer-Integrated Manufacturing},
  volume={79},
  pages={102432},
  year={2023},
  publisher={Elsevier}
}

@article{tolgyessy2021skeleton,
  title={Skeleton tracking accuracy and precision evaluation of Kinect V1, Kinect V2, and the azure kinect},
  author={T{\"o}lgyessy, Michal and Dekan, Martin and Chovanec, L'ubo{\v{s}}},
  journal={Applied Sciences},
  volume={11},
  number={12},
  pages={5756},
  year={2021},
  publisher={Multidisciplinary Digital Publishing Institute}
}

\end{document}